\documentclass{article}
\usepackage[preprint]{tmlr}                 
\usepackage{amsmath,amssymb}
\usepackage{booktabs}
\usepackage{graphicx}
\usepackage{microtype}

\title{When Is an LLM Worth It for Hyperparameter Optimization?\\
A Budget-Matched Study on Tabular Data Finds the Warm-Start Is a Default Configuration, Not the Model}

\author{\name Carson Rodrigues \email carson@celabe.com \\
        \addr Celabe
        \AND
        \name Oysturn Vas \email ovas@uwaterloo.ca \\
        \addr University of Waterloo
        \AND
        \name Isaiah Abner DCosta \email i.dcosta@student.uq.edu.au \\
        \addr University of Queensland (EAIT)
        \AND
        \name Nithish Kumar Prabhakaran \email nithishkumar.prabhakaran@student.uq.edu.au \\
        \addr University of Queensland (EAIT)}

\def\month{06}\def\year{2026}\def\openreview{\url{https://openreview.net/forum?id=XXXX}}

\begin{document}
\maketitle

\begin{abstract}
Large language models (LLMs) have been proposed as hyperparameter-optimization (HPO) advisors that ``warm-start'' search from prior knowledge, proposing strong configurations in very few evaluations. We test that warm-start claim under a budget-matched, multi-seed protocol on eight PMLB tabular benchmarks, comparing an LLM advisor (LLM-OptFlow) against four classical baselines (random search, a full Optuna-TPE study, Gaussian-process Bayesian optimization [GP-BO], and successive halving) over one shared search space, with paired tests and bootstrap 95\% confidence intervals across the $8\times5=40$ (task, seed) units. Our central finding is cautionary. The advisor's strong first point is not an LLM output at all: like prior LLM-HPO systems the loop is seeded with a fixed default configuration, evaluated before any model call, which alone reaches $88.7\%$ mean best-CV, identical to within $0.01$\,pp across all seven advisor models we test. The LLM's own proposals add only $+0.40$ percentage points (pp) of cross-validation accuracy over that seed across the 12-evaluation budget and \emph{nothing} on held-out test (LLM$-$Default $=-0.01$\,pp, $p=0.92$). When the same default seed is granted to classical search, the advisor's apparent lead collapses: against seeded random search (an exact control) it leads by only $+0.20$\,pp at a 2-evaluation budget, the lead is gone by 5 evaluations, and it is \emph{behind} by 12 ($-0.37$\,pp); seeding the model-based optimizers too only widens this. Even without that seed, classical search ties the advisor by 12 evaluations and passes it by 40 ($+0.6$ to $+0.8$\,pp, $p\le10^{-4}$). Two LLM-specific behaviors do survive: a single-task exploration failure (\texttt{vehicle}, where the advisor stays at the default while classical search gains $6$--$9$\,pp; random $+9.06$\,pp, $p<0.01$) and a rule-based confidence filter that is operationally useful (it removes out-of-space proposals and ${\sim}33\%$ of wasted compute) but does not change accuracy. The recommendation is deflationary: on tabular HPO, seed classical search with a sensible default; an LLM advisor adds no measurable generalization benefit and is overtaken within a handful of evaluations. We release the harness and a script that reproduces every statistic.
\end{abstract}

\section{Introduction}
Hyperparameter optimization (HPO) is a core, expensive step in applied machine learning. Classical methods are well understood but treat the search as black-box: random search \citep{bergstra2012random}, Bayesian optimization \citep{snoek2012practical}, Tree-structured Parzen Estimators (TPE) as in Optuna \citep{optuna_2019}, and multi-fidelity methods such as Hyperband \citep{li2017hyperband}. A recent line of work instead prompts an LLM to propose configurations from a natural-language description of the task and the history of attempts \citep{yang2024opro,liu2024agenthpo,mahammadli2024sllmbo,liu2024llambo,liu2023llm_hp}. The appeal is prior knowledge: an LLM ``knows'' that gradient boosting often beats logistic regression on tabular data, and is claimed to propose a strong starting configuration in very few evaluations, in effect a learned warm-start. In practice these advisor loops are also \emph{seeded} with a fixed default configuration to begin the search, a detail that turns out to be decisive.

The question for a practitioner is not only whether LLM advising ever helps, but \emph{what} provides the help. Prior empirical claims of LLM superiority were undermined by under-powered baselines (e.g., three-trial proxies for tuned search), single runs, and no statistics; they also credited the model with a warm-start without separating its proposals from the default configuration the loop is initialized with. We pull these apart under matched budgets with proper variance estimates, and find that on tabular HPO the warm-start is the default seed, not the model: the LLM's own proposals add at most $0.4$ percentage points on the cross-validation objective and nothing on held-out accuracy, and once classical search is given the same default it matches the advisor within a few evaluations and surpasses it thereafter.

\paragraph{Contributions.}
\begin{itemize}\itemsep1pt
\item A \textbf{budget-matched, multi-seed} comparison of an LLM advisor against four classical baselines over a single shared search space on eight standard PMLB benchmarks \citep{olson2017pmlb}: random search, a \emph{full} Optuna-TPE study, Gaussian-process Bayesian optimization \citep{snoek2012practical}, and successive halving \citep{li2017hyperband}, with paired tests and bootstrap 95\% CIs \citep{efron1986bootstrap}. The decisive addition is a \textbf{control that seeds the classical baselines with the same default} the advisor is initialized with, which isolates the LLM's marginal contribution.
\item The central, \textbf{cautionary finding: the warm-start is a default configuration, not the model}. The advisor's first point is a fixed default seed (model-independent, $88.7\%$ mean best-CV); the LLM's own proposals add only $+0.40$\,pp of CV accuracy over that seed and $-0.01$\,pp on held-out test ($p=0.92$); and once the same default is given to classical search, the advisor's edge is $+0.20$\,pp at 2 evaluations (n.s.\ vs.\ TPE), gone by 5, and negative by 12.
\item A \textbf{negative practical recommendation}: on tabular HPO, default-seeded classical search matches the LLM advisor within a few evaluations and surpasses it thereafter; the advisor also ties un-seeded classical search by 12 evaluations and is passed by 40, at a far lower cost per proposal and no API dependency.
\item Two \textbf{surviving LLM-specific phenomena}, each characterized and honestly bounded: a single-task exploration failure (\texttt{vehicle}) and a rule-based confidence filter whose value is operational, not accuracy-improving.
\item A \textbf{seven-model panel} (five providers, nano-to-frontier) confirming the default-driven first point is model-independent, and a released, reproducible harness with a script that regenerates every statistic.
\end{itemize}

\section{Related Work}
\paragraph{Classical HPO.} Random search is a strong, embarrassingly-parallel baseline \citep{bergstra2012random}; model-based methods (GP-BO \citep{snoek2012practical}, SMAC \citep{hutter2011smac}, TPE \citep{optuna_2019}) improve sample efficiency, and multi-fidelity methods \citep{li2017hyperband} improve anytime performance. Toolkits such as Optuna \citep{optuna_2019} and KerasTuner \citep{omalley2019kerastuner} make these routine.

\paragraph{Warm-starting and meta-learning for HPO.} Initializing search from configurations that performed well on related tasks is a long-studied route to strong early evaluations: meta-learning the initial design for Bayesian optimization \citep{feurer2015metainit} and the warm-start in auto-sklearn \citep{feurer2015autosklearn} are direct precedents. An LLM advisor's prior knowledge is, in this framing, a learned warm-start, so a fair evaluation must separate it from the trivial warm-start of a hand-chosen default configuration. Supplying that control is central to our study.

\paragraph{LLMs for optimization and HPO.} OPRO frames the LLM itself as the optimizer over a prompt of prior (solution, score) pairs \citep{yang2024opro}. A complementary thread optimizes the \emph{textual} components of a system rather than its numeric configuration: GEPA evolves prompts, code, and control flow by having an LLM reflect in natural language on full execution traces and mutate candidates on a Pareto front, reportedly beating reinforcement learning (GRPO) while using up to $35\times$ fewer rollouts \citep{agrawal2026gepa}. Our object of optimization differs (numeric model and hyperparameter configurations under matched evaluation budgets), but the same question drives both lines of work: when does reflective LLM proposal beat brute-force search? AgentHPO \citep{liu2024agenthpo} and SLLMBO \citep{mahammadli2024sllmbo} build iterative LLM HPO loops, the latter hybridizing with TPE; LLAMBO \citep{liu2024llambo} uses LLMs inside Bayesian optimization, including for warm-starting. Recent evidence is converging on the limits of LLM proposers: a reproducibility study of LLM Bayesian optimization finds the LLM surrogate weaker than Gaussian-process and SMAC search \citep{rychert2025llambo_repro}, a high-dimensional study reports LLMs competitive only below roughly a dozen features and overtaken by Bayesian search beyond that \citep{srinivasan2026beyond}, and a re-examination of OPRO documents small-scale LLMs failing as optimizers \citep{zhang2024revisitopro}; a parallel proposal drops the classical baseline altogether \citep{naphade2025smallexpert}, the very comparison our default-seed control restores. Agentic ML-engineering benchmarks (MLAgentBench \citep{huang2024mlagentbench}, MLE-bench \citep{chan2025mlebench}) evaluate broader pipelines. Our contribution is not a new method but a \emph{rigorous characterization} of what, if anything, the LLM-advisor pattern buys, which prior work asserts more than measures. In contrast to claims of LLM sample-efficiency, we find that on tabular HPO the early advantage is attributable to the default seed rather than the model's proposals, a control these papers do not isolate.

\section{Method: LLM-OptFlow and the Confidence Filter}
\label{sec:method}
\textbf{Search space $S$.} All methods explore one space over four model families (logistic regression, random forest, gradient boosting, and SVC) with standard hyperparameter ranges (e.g., RF \texttt{n\_estimators}$\in[50,500]$, GB \texttt{learning\_rate}$\in[10^{-3},0.3]$). Random search samples $S$ uniformly; TPE explores the identical space; the LLM is prompted with the ranges and asked to stay within them. All methods thus explore the same space and objective; \S\ref{sec:results} also controls for the default configuration the advisor is seeded with, by granting the same seed to classical search.

\textbf{LLM-OptFlow advisor.} The loop is \emph{seeded} with a fixed default configuration (\texttt{RandomForest} with \texttt{n\_estimators}$=100$, \texttt{max\_depth}$=16$, \texttt{min\_samples\_leaf}$=1$; the same configuration as the \texttt{Default} baseline), scored once \emph{before any LLM call}; this seed is the first point on the advisor's budget curve. Thereafter, at each iteration the LLM (Claude Haiku) receives the dataset description, the incumbent configuration and its 3-fold CV score, and the history of attempts, and proposes one configuration with a short rationale. The objective is 3-fold CV accuracy on the training split; the reported number is held-out test accuracy of the selected configuration. We count the default seed as the advisor's first evaluation, so its $k$-th budget point reflects the seed plus $k-1$ LLM proposals.

\textbf{Confidence filter.} Each proposal is validated against $S$ before evaluation: unknown models, out-of-range numeric values, bad enums, or wrong types are rejected and the incumbent is kept (no evaluation spent). A naive validator that checks only trivial bounds never fires in practice; ours enforces the full space $S$, so it triggers exactly when a proposal drifts outside $S$.

\section{Experimental Setup}
\textbf{Benchmarks.} Eight PMLB \citep{olson2017pmlb} classification datasets spanning sizes and class counts: credit-g, spambase, phoneme, churn, satimage (6-class), vehicle (4-class), ionosphere, hypothyroid. (OpenML's live API was unavailable during our study; PMLB provides the same canonical datasets from a reliable mirror.)

\textbf{Protocol.} For each dataset and each of 5 seeds we take a stratified 70/30 train/test split. Random search, TPE, and GP-BO run 40 trials; LLM-OptFlow runs 12 iterations. We log the running-best CV score after each evaluation (the budget curve) and report test accuracy of the final selected configuration. Every number is reported as a mean$\pm$std over the five seeds, and the budget curves show standard-error bands. Optuna's TPE and GP-BO samplers use the default 10 random startup trials before their model-based phase, so their surrogate machinery does not engage until trial~11 and the single-digit-budget regime reflects each sampler's startup behavior rather than its surrogate.

\textbf{Statistical analysis.} All cross-method comparisons are \emph{paired} across the $8\times5=40$ (task, seed) units. We report mean differences in percentage points (pp) with bootstrap 95\% confidence intervals (20{,}000 resamples) \citep{efron1986bootstrap} and paired two-sided tests (Student's $t$, with the Wilcoxon signed-rank test as a distribution-free check). A released script (\texttt{significance.py}) regenerates every statistic reported below.

\textbf{Strong classical baselines.} Beyond random search and TPE we add two stronger classical optimizers, all exploring the identical space $S$ and objective. \emph{GP-BO} is Gaussian-process Bayesian optimization \citep{snoek2012practical} via Optuna's GPSampler; because GP-BO assumes a fixed-dimensional space, we give it the standard flat encoding of $S$ (all hyperparameters present every trial, those irrelevant to the sampled family masked at evaluation), so it is full-fidelity and directly comparable on the per-evaluation budget curve. \emph{Successive halving} (SH) is a Hyperband-family multi-fidelity method \citep{li2017hyperband} (scikit-learn's \texttt{HalvingRandomSearchCV}) using training-set size as the fidelity; being multi-fidelity it spends partial-fidelity evaluations, so we do \emph{not} place it on the per-evaluation budget curve (that accounting would be unfair) and instead report it in the final-budget table with its total resource noted. SMAC's random-forest surrogate would be a natural further baseline; its native build was unavailable on our platform, and GP-BO and TPE together already span the GP- and tree-based model-based-BO families.

\section{Results}
\label{sec:results}

\subsection{The warm-start is a default configuration, not the model}
\label{ssec:warmstart}
\begin{figure}[t]\centering
\includegraphics[width=0.7\textwidth]{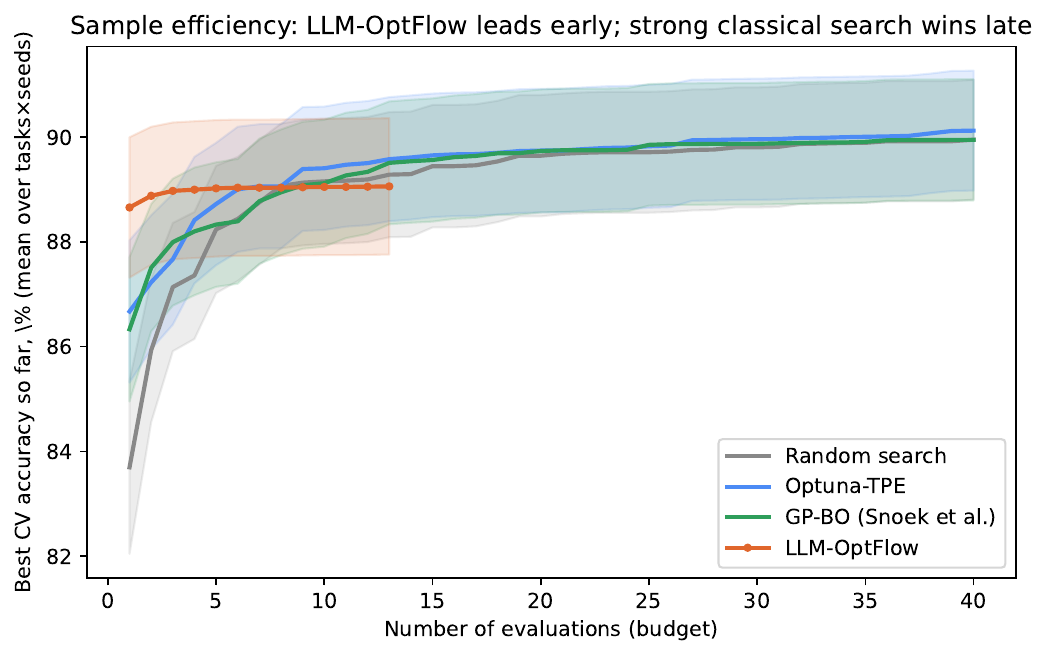}
\caption{Running-best CV accuracy vs.\ evaluation budget (mean $\pm$ s.e.\ over tasks$\times$seeds). LLM-OptFlow is seeded with a fixed default (its budget-1 point), so its early lead reflects that seed; once classical search is given the same seed the advantage disappears, and even un-seeded, random, TPE, and GP-BO (40 evals) tie the advisor by 12 evaluations and pass it by 40.}
\label{fig:budget}
\end{figure}

Table~\ref{tab:budget} and Figure~\ref{fig:budget} report mean best-CV accuracy at matched budgets. The advisor's first point is striking: $88.7\%$ at one evaluation, far above random search ($83.7\%$), TPE ($86.7\%$), and GP-BO ($86.3\%$). But it is \emph{not} an LLM proposal. It is the fixed default configuration the loop is seeded with, scored before any model is queried (\S\ref{sec:method}); across all seven advisors we test it is identical to within $0.01$\,pp (Table~\ref{tab:multimodel}; the residual reflects cross-validation nondeterminism, not the model), as it essentially must be, since the model has not yet been called. The \texttt{Default (fixed)} row in Table~\ref{tab:budget} is this same configuration, flat across budget. The classical baselines receive no such seed, so the apparent ``$+2$ to $+5$\,pp warm-start'' at one evaluation is a hand-chosen default beating a single cold draw, not the LLM's prior knowledge.

\textbf{What the LLM actually adds.} Measured over its own seed, the LLM's first proposal improves best-CV by only $+0.22$\,pp (95\% CI $[0.09,0.42]$, $p=0.02$) and its proposals over the full 12-evaluation budget by $+0.40$\,pp $[0.22,0.62]$ ($p<0.001$): a real but small effect on the CV objective. On held-out test accuracy it adds nothing: LLM-OptFlow is statistically indistinguishable from simply using the default ($-0.01$\,pp, 95\% CI $[-0.22,0.19]$, $p=0.92$).

\textbf{The decisive control: give classical search the same seed.} Seeding random search with the same default (an exact control, since random draws are independent of the seed), the advisor leads by only $+0.20$\,pp at a 2-evaluation budget ($p=0.03$), the lead is gone by 5 evaluations ($-0.09$\,pp, n.s.), and the advisor is \emph{behind} by 12 ($-0.37$\,pp, 95\% CI $[-0.82,-0.02]$). An approximate control that also seeds the model-based optimizers (treating the default as a performance floor; see Limitations) tells the same story: a non-significant $+0.14$\,pp against TPE at 2 evaluations, and $-0.60$/$-0.48$\,pp behind TPE/GP-BO by 12. Once the free default is available to both, the LLM's proposals buy no sample efficiency.

\textbf{Even un-seeded, classical search catches up.} By 12 evaluations the advisor is statistically tied with both Bayesian optimizers without any seed of their own (TPE$-$LLM $=+0.45$\,pp, 95\% CI $[-0.03,+1.01]$, paired-$t$ $p=0.10$, Wilcoxon $p=0.81$; GP-BO$-$LLM $=+0.28$\,pp $[-0.22,+0.85]$, $p=0.31$). It then falls behind because classical search keeps improving (TPE $+0.62$\,pp, random $+0.76$\,pp, GP-BO $+0.61$\,pp from 12 to 40 evals, all $p\le10^{-4}$) while the advisor has converged: its 5-to-12 gain is $0.03$\,pp and extending it to the full 40-evaluation budget on four datasets adds at most $0.24$\,pp (credit-g $+0.24$, spambase $+0.21$, satimage $+0.00$, vehicle $+0.00$). The advisor's ceiling is set early and low, and that both a tree-based (TPE) and a GP-based (GP-BO) optimizer overtake it indicates the effect is a property of classical search in general, not of one sampler.

\begin{table}[t]\centering
\caption{Mean best-CV accuracy (\%) at matched evaluation budgets (8 tasks $\times$ 5 seeds). \emph{Default (fixed)} is a single hand-chosen configuration, constant across budget. LLM-OptFlow is \emph{seeded} with it, so its 1-evaluation entry ($^\dagger$) \emph{is} the default ($88.7$), not an LLM proposal; the LLM's proposals add $\le0.4$\,pp over this seed (\S\ref{ssec:warmstart}). Once the same seed is given to classical search the advisor's early lead disappears; even un-seeded, classical search ties the advisor by 12 evaluations and passes it by 40. Bold marks the best tuned, non-seeded optimizer at the 12- and 40-evaluation budgets; the advisor's seed-driven early lead is not bolded.}
\label{tab:budget}
\begin{tabular}{lccccc}
\toprule
Budget (evals) & 1 & 3 & 5 & 12 & 40 \\
\midrule
Default (fixed) & 88.7 & 88.7 & 88.7 & 88.7 & 88.7 \\
Random search   & 83.7 & 87.1 & 88.2 & 89.2 & 90.0 \\
Optuna-TPE       & 86.7 & 87.7 & 88.7 & \textbf{89.5} & \textbf{90.1} \\
GP-BO            & 86.3 & 88.0 & 88.3 & 89.3 & 89.9 \\
LLM-OptFlow      & 88.7$^\dagger$ & 89.0 & 89.0 & 89.1 & --- \\
\bottomrule
\end{tabular}
\end{table}

\subsection{Final-budget accuracy and a failure mode}
Table~\ref{tab:final} gives held-out test accuracy at full budget. On most datasets all methods, including a sensible default, agree within ${\sim}1$ point: on the seven non-\texttt{vehicle} tasks the advisor and TPE are statistically indistinguishable (TPE$-$LLM $=+0.20$\,pp, 95\% CI $[-0.01,+0.43]$, paired-$t$ $p=0.10$, $n=35$), confirming that with adequate trials these tabular problems are not separated by the optimizer. The informative exception is \texttt{vehicle}: random search, TPE, and GP-BO all find a substantially better configuration ($82.4$/$81.4$/$79.8\%$), and even successive halving reaches $79.3\%$, while LLM-OptFlow remains essentially at the default ($73.3\%$ vs.\ default $73.5\%$; per method, random$-$LLM $=+9.06$, TPE $+8.11$, GP-BO $+6.54$, SH $+5.98$\,pp, with $p<0.01$ for random/TPE/GP-BO and $p=0.02$ for SH; random$-$LLM 95\% CI $[5.59,12.13]$). This one task in fact accounts for the entire pooled test-accuracy gap between classical search and the advisor (TPE$-$LLM over all eight datasets $=+1.18$\,pp, $p=0.01$, but a non-significant $+0.20$\,pp once \texttt{vehicle} is removed); on every other task the advisor, the default, and tuned classical search are mutually indistinguishable. That \emph{every} classical optimizer, including model-based Bayesian optimization, escapes this basin while the LLM does not isolates the failure as one of exploration, not of the search budget or the sampler: the LLM repeatedly proposes ``reasonable'' configurations near its prior, and even given the full 40-evaluation budget it never escapes ($76.1\%$ best-CV at both 12 and 40 evaluations).

\begin{table}[t]\centering\footnotesize\setlength{\tabcolsep}{4pt}
\caption{Held-out test accuracy (\%, mean$\pm$std over 5 seeds) at full budget. Random/TPE/GP-BO: 40 evals; LLM-OptFlow: 12; successive halving (SH) is multi-fidelity, consuming ${\sim}6$ full-fidelity-equivalent evals. On \texttt{vehicle}, the one task the optimizer separates, \emph{every} classical method beats both the default and the LLM (bold marks the best classical optimizer there).}
\label{tab:final}
\begin{tabular}{lcccccc}
\toprule
Dataset & Default & Random & Optuna-TPE & GP-BO & SH & LLM-OptFlow \\
\midrule
churn       & 95.4$\pm$0.5 & 95.5$\pm$0.5 & 95.5$\pm$0.5 & 95.6$\pm$0.5 & 95.3$\pm$0.4 & 95.3$\pm$0.4 \\
credit-g    & 76.4$\pm$0.6 & 75.5$\pm$1.6 & 77.1$\pm$0.9 & 76.8$\pm$1.0 & 74.7$\pm$2.9 & 76.6$\pm$0.9 \\
hypothyroid & 98.1$\pm$0.1 & 98.1$\pm$0.2 & 98.1$\pm$0.3 & 98.0$\pm$0.5 & 97.8$\pm$0.3 & 98.0$\pm$0.4 \\
ionosphere  & 93.6$\pm$2.2 & 93.6$\pm$2.8 & 93.6$\pm$2.2 & 93.8$\pm$2.2 & 89.4$\pm$3.2 & 93.4$\pm$2.7 \\
phoneme     & 90.2$\pm$0.5 & 90.2$\pm$0.8 & 90.4$\pm$0.6 & 90.3$\pm$0.7 & 89.8$\pm$0.9 & 90.2$\pm$0.5 \\
satimage    & 91.2$\pm$0.6 & 91.4$\pm$0.5 & 91.3$\pm$0.6 & 91.5$\pm$0.6 & 89.3$\pm$1.9 & 91.3$\pm$0.5 \\
spambase    & 95.0$\pm$0.3 & 95.3$\pm$0.7 & 95.5$\pm$0.6 & 95.4$\pm$0.5 & 93.9$\pm$1.3 & 95.1$\pm$0.5 \\
vehicle     & 73.5$\pm$0.9 & \textbf{82.4$\pm$2.7} & 81.4$\pm$2.0 & 79.8$\pm$1.3 & 79.3$\pm$2.5 & 73.3$\pm$1.5 \\
\bottomrule
\end{tabular}
\end{table}

\subsection{The confidence filter, finally exercised}
\begin{figure}[t]\centering
\includegraphics[width=0.7\textwidth]{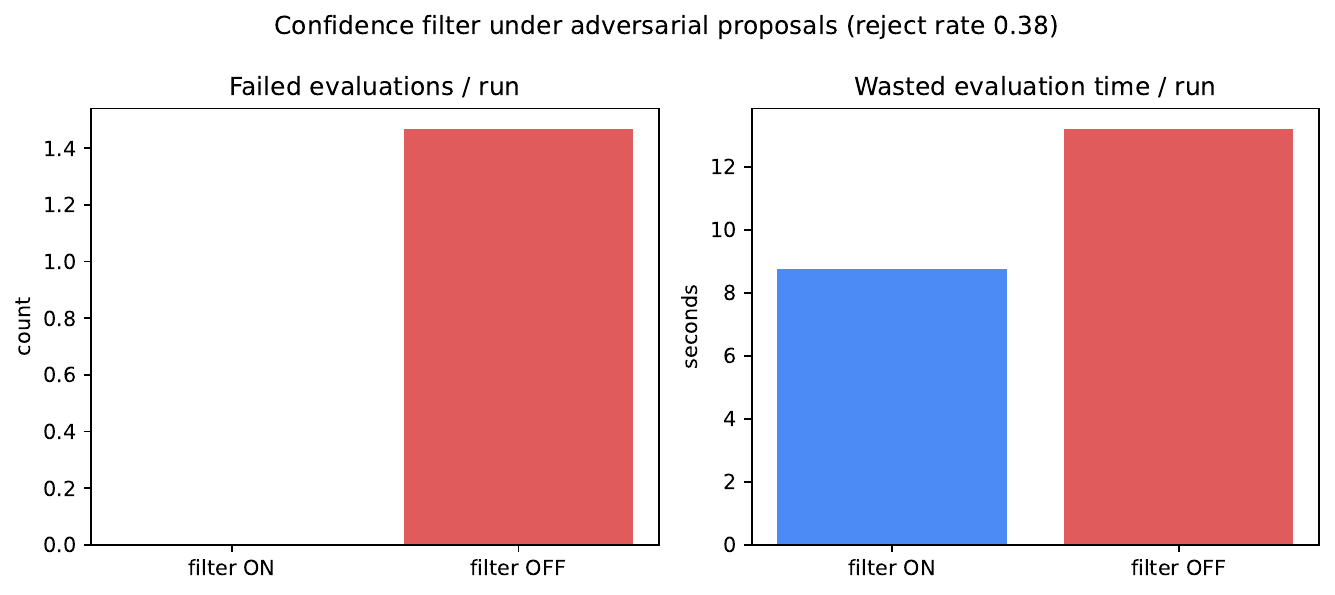}
\caption{Confidence filter under adversarial proposals: failed evaluations and wasted evaluation time per run, filter ON vs.\ OFF.}
\label{fig:conf}
\end{figure}

On well-formed proposals the filter never fires (rejection rate $0.00$). To test its value, on three representative datasets (credit-g, spambase, churn) we inject an adversarial proposal channel that corrupts a fraction of proposals into configurations outside $S$ (unknown model, out-of-range value, wrong type), the drift LLMs actually produce. Under a $0.40$ corruption rate the filter rejects at $0.38$ (near-perfect precision/recall, since it is rule-based), eliminating failed evaluations ($1.5\to0$ per run) and cutting wasted evaluation time by ${\sim}33\%$ ($13.2\to8.8$\,s), while final accuracy is unchanged ($0.889$ vs.\ $0.890$); see Figure~\ref{fig:conf}. In a keep-best loop the filter does not improve accuracy, since bad configurations simply score poorly and are never selected, so its value is \emph{operational}: it prevents wasted compute and runtime failures in an LLM-in-the-loop system.

\subsection{Sensitivity to the LLM: a seven-model panel}
\begin{figure}[t]\centering
\includegraphics[width=0.7\textwidth]{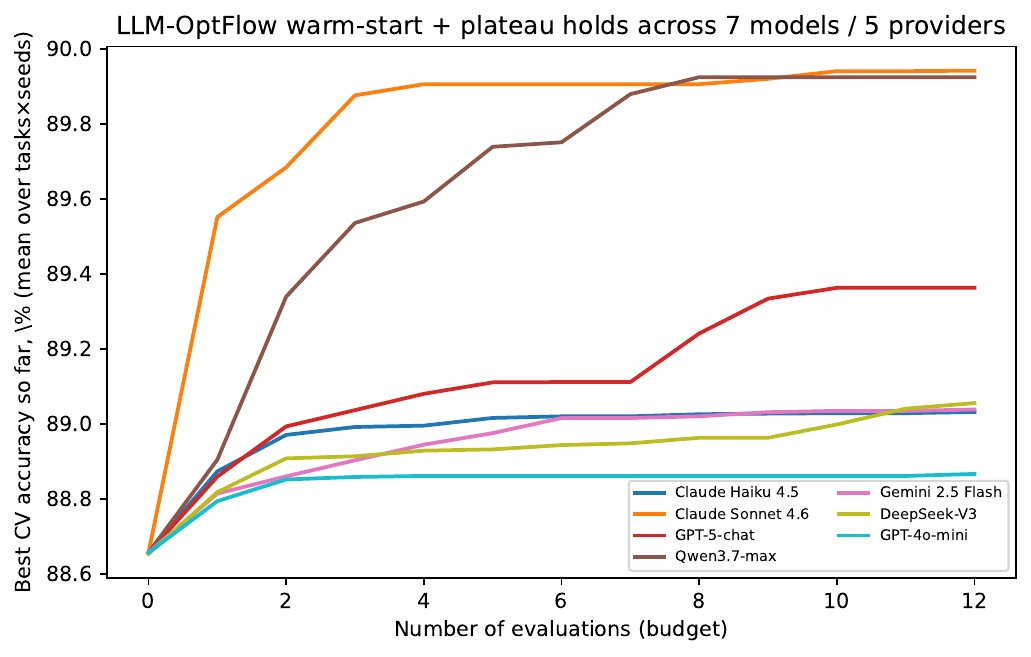}
\caption{Running-best CV accuracy vs.\ budget for all seven advisor models (mean over tasks$\times$seeds). The budget-1 point is the shared default (identical across models); the LLM proposals plateau quickly across every model and provider, with frontier models plateauing slightly higher.}
\label{fig:multimodel}
\end{figure}

To test whether our findings depend on the specific advisor, we re-ran the identical advisor, confidence filter, and search space $S$ across \emph{seven} models spanning five providers and the nano-to-frontier range (Claude Haiku 4.5 and Sonnet 4.6, GPT-5-chat and GPT-4o-mini, Gemini 2.5 Flash, DeepSeek-V3, and Qwen3.7-max), through one OpenRouter pipeline,\footnote{Exact OpenRouter model identifiers, queried June 2026: \texttt{anthropic/claude-haiku-4.5}, \texttt{anthropic/claude-sonnet-4.6}, \texttt{openai/gpt-5-chat}, \texttt{openai/gpt-4o-mini}, \texttt{google/gemini-2.5-flash}, \texttt{deepseek/deepseek-chat} (DeepSeek-V3 series), and \texttt{qwen/qwen3.7-max}. Provider model slugs may be deprecated or re-versioned over time; the released harness records the identifiers used.} on the same 8 datasets / 5 seeds / 12-iteration budget (Table~\ref{tab:multimodel}, Figure~\ref{fig:multimodel}). Three findings emerge.

First, the \textbf{first point is the seed, identical across models}. Every advisor's budget-1 value is the same $88.7\%$ default (Table~\ref{tab:budget}), confirming it is not an LLM output. The first genuine LLM \emph{proposal} (Table~\ref{tab:multimodel}, CV@1) lands at $88.8$--$89.6\%$; but at the matched 2-evaluation budget classical search has already reached random $85.9\%$, TPE $87.2\%$, GP-BO $87.5\%$, so after one real proposal the advisor's edge ranges from a fraction of a point (cheaper models) to ${\sim}2$\,pp (frontier), consistent with \S\ref{ssec:warmstart}.

Second, \textbf{final accuracy varies across models, but only one task separates them at all}. Just one dataset (\texttt{vehicle}) separates the optimizers, so this observation rests on that task. There, the two strongest advisors, \emph{Sonnet 4.6} and \emph{Qwen3.7-max}, reach the classical $82\%$ and lead the panel ($90.3\%$ final test accuracy vs.\ $89.2$--$89.5\%$ for the rest), while the others stay near the default. The relation is non-monotone, however: GPT-5-chat, also a frontier model, does \emph{not} escape ($75.1\%$), and once \texttt{vehicle} is removed all seven models tie (final ${\sim}91.5\%$). We therefore report this as a single-task observation, not a scaling law.

Third, the \texttt{vehicle} \textbf{exploration outcome is model-dependent on this one task}. Sonnet 4.6 and Qwen3.7-max escape the default basin ($82.0\pm2.0\%$, above the default on all five seeds, matching classical search's $82.4\%$), while the other five, including GPT-5-chat ($75.1\%$), remain near the default ($73$--$75\%$). Across all seven models the confidence filter fired at essentially zero on well-formed proposals ($\le0.002$), and the panel's Haiku 4.5 run reproduces the canonical Haiku result per seed to within run-to-run noise, confirming the two pipelines are consistent.

\begin{table}[t]\centering\footnotesize\setlength{\tabcolsep}{4pt}
\caption{Seven-model panel (8 datasets $\times$ 5 seeds, 12-iteration budget, one OpenRouter pipeline). CV@$k$ is mean best-CV (\%) after $k$ LLM \emph{proposals}; the fixed default seed is proposal~0, a uniform $88.7\%$ across all models (Table~\ref{tab:budget}). This proposal index is offset by one from Table~\ref{tab:budget}'s per-evaluation axis, which counts the seed as evaluation~1. Final and \texttt{vehicle} are mean held-out test accuracy (\%); reject is the confidence-filter rate. The seed-driven start is model-independent; final accuracy and \texttt{vehicle} escape vary with model, but only \texttt{vehicle} separates the optimizers (rows sorted by final accuracy).}
\label{tab:multimodel}
\begin{tabular}{lcccccc}
\toprule
Model & CV@1 & CV@6 & CV@12 & Final & \texttt{vehicle} & reject \\
\midrule
Claude Sonnet 4.6 & 89.6 & 89.9 & 89.9 & \textbf{90.3} & \textbf{82.0} & 0.00 \\
Qwen3.7-max       & 88.9 & 89.8 & 89.9 & \textbf{90.3} & \textbf{82.0} & 0.00 \\
GPT-5-chat        & 88.9 & 89.1 & 89.4 & 89.5 & 75.1 & 0.00 \\
Claude Haiku 4.5  & 88.9 & 89.0 & 89.0 & 89.2 & 73.4 & 0.00 \\
Gemini 2.5 Flash  & 88.8 & 89.0 & 89.0 & 89.3 & 73.7 & 0.00 \\
DeepSeek-V3       & 88.8 & 88.9 & 89.1 & 89.3 & 73.2 & 0.00 \\
GPT-4o-mini       & 88.8 & 88.9 & 88.9 & 89.3 & 73.4 & 0.00 \\
\bottomrule
\end{tabular}
\end{table}

\section{Discussion}
The evidence supports a cautionary, practical claim: \textbf{on tabular HPO, the warm-start attributed to LLM advisors is a default configuration, not the model.} A single hand-chosen default reaches the advisor's headline first-evaluation accuracy; the LLM's proposals add $\le0.4$\,pp on the CV objective and nothing on held-out test ($p=0.92$); and once the same default is granted to classical search, the advisor's early lead is gone within five evaluations and reversed by twelve. Even without the seed, tuned classical search ties the advisor by twelve evaluations and passes it by forty, at a far lower cost per proposal and with no API dependency. The actionable recommendation is therefore the opposite of the usual pitch: seed classical search with a sensible default, or a cheap meta-learned initialization \citep{feurer2015metainit}, rather than pay for an LLM in the loop. The two LLM-specific behaviors we did find are a caution and a convenience, not a case for the advisor: the \texttt{vehicle} exploration failure shows the advisor can lock onto its prior and miss a basin every classical method finds, and the one component that helped, the confidence filter, is a rule-based validator that needs no LLM at all.

\section{Limitations}
Our seeded-baseline control is \emph{exact} for random search (random draws are independent of the seed) but approximate for the model-based optimizers: enqueuing the default would also shift TPE's and GP-BO's subsequent draws, so we report the exact random-search control and note that un-seeded TPE and GP-BO already tie and then pass the advisor (\S\ref{ssec:warmstart}), so seeding them could only strengthen the conclusion; a fully re-seeded model-based study is straightforward future work. The capability-related observations rest on a single exploration-separating task (\texttt{vehicle}); confirming any scaling relationship needs more exploration-hard datasets. We study tabular classification with sklearn model families; deep-learning HPO (where tiny budgets, and thus a good initialization, matter most) is the natural extension, but its per-trial cost made a budget-matched, multi-seed study impractical on our hardware. Our seven-model panel covers five providers and the nano-to-frontier range, though all are general-purpose chat models rather than HPO-specialized agents. We run many paired tests without a multiple-comparison correction; the load-bearing results survive a Bonferroni adjustment ($p\le10^{-4}$ throughout), and the one borderline result (pooled TPE$-$LLM, $p=0.015$) is used only to \emph{deflate} the classical advantage, so correction would only reinforce our conclusions. The adversarial channel is synthetic but mimics observed LLM drift.

\section{Conclusion}
Under a rigorous, budget-matched, multi-seed protocol, the apparent sample-efficiency of an LLM hyperparameter advisor on tabular data is an artifact of the default configuration it is seeded with: the model's own proposals add at most $0.4$\,pp on the cross-validation objective and no measurable held-out improvement, and default-seeded classical search matches the advisor within a few evaluations and surpasses it thereafter. The honest recommendation is to seed classical search with a good default rather than pay for an LLM in the loop. We release the harness and a script that reproduces every statistic, so both the claim and its limits can be checked directly.

\bibliographystyle{tmlr}
\bibliography{references}

\begin{thebibliography}{22}
\providecommand{\natexlab}[1]{#1}
\providecommand{\url}[1]{\texttt{#1}}
\expandafter\ifx\csname urlstyle\endcsname\relax
  \providecommand{\doi}[1]{doi: #1}\else
  \providecommand{\doi}{doi: \begingroup \urlstyle{rm}\Url}\fi

\bibitem[Agrawal et~al.(2026)Agrawal, Tan, Soylu, Ziems, Khare, Opsahl-Ong,
  Singhvi, Shandilya, Ryan, Jiang, Potts, Sen, Dimakis, Stoica, Klein, Zaharia,
  and Khattab]{agrawal2026gepa}
Lakshya~A. Agrawal, Shangyin Tan, Dilara Soylu, Noah Ziems, Rishi Khare, Krista
  Opsahl-Ong, Arnav Singhvi, Herumb Shandilya, Michael~J. Ryan, Meng Jiang,
  Christopher Potts, Koushik Sen, Alexandros~G. Dimakis, Ion Stoica, Dan Klein,
  Matei Zaharia, and Omar Khattab.
\newblock {GEPA}: Reflective prompt evolution can outperform reinforcement
  learning.
\newblock In \emph{International Conference on Learning Representations
  (ICLR)}, 2026.
\newblock Oral presentation; arXiv:2507.19457.

\bibitem[Akiba et~al.(2019)Akiba, Sano, Yanase, Ohta, and Koyama]{optuna_2019}
Takuya Akiba, Shotaro Sano, Toshihiro Yanase, Takeru Ohta, and Masanori Koyama.
\newblock Optuna: A next-generation hyperparameter optimization framework.
\newblock In \emph{Proceedings of the 25th ACM SIGKDD International Conference
  on Knowledge Discovery and Data Mining}, pp.\  2623--2631, 2019.
\newblock \doi{10.1145/3292500.3330701}.

\bibitem[Bergstra \& Bengio(2012)Bergstra and Bengio]{bergstra2012random}
James Bergstra and Yoshua Bengio.
\newblock Random search for hyper-parameter optimization.
\newblock \emph{Journal of Machine Learning Research}, 13:\penalty0 281--305,
  2012.

\bibitem[Chan et~al.(2025)]{chan2025mlebench}
Jun~Shern Chan et~al.
\newblock Mle-bench: Evaluating machine learning agents on machine learning
  engineering.
\newblock \emph{International Conference on Learning Representations (ICLR)},
  2025.

\bibitem[Efron \& Tibshirani(1986)Efron and Tibshirani]{efron1986bootstrap}
Bradley Efron and Robert Tibshirani.
\newblock Bootstrap methods for standard errors, confidence intervals, and
  other measures of statistical accuracy.
\newblock \emph{Statistical Science}, 1\penalty0 (1):\penalty0 54--75, 1986.

\bibitem[Feurer et~al.(2015{\natexlab{a}})Feurer, Klein, Eggensperger,
  Springenberg, Blum, and Hutter]{feurer2015autosklearn}
Matthias Feurer, Aaron Klein, Katharina Eggensperger, Jost~Tobias Springenberg,
  Manuel Blum, and Frank Hutter.
\newblock Efficient and robust automated machine learning.
\newblock In \emph{Advances in Neural Information Processing Systems},
  volume~28, 2015{\natexlab{a}}.

\bibitem[Feurer et~al.(2015{\natexlab{b}})Feurer, Springenberg, and
  Hutter]{feurer2015metainit}
Matthias Feurer, Jost~Tobias Springenberg, and Frank Hutter.
\newblock Initializing bayesian hyperparameter optimization via meta-learning.
\newblock In \emph{Proceedings of the AAAI Conference on Artificial
  Intelligence}, volume~29, 2015{\natexlab{b}}.

\bibitem[Huang et~al.(2024)Huang, Vora, Liang, and
  Leskovec]{huang2024mlagentbench}
Qian Huang, Jian Vora, Percy Liang, and Jure Leskovec.
\newblock Mlagentbench: Evaluating language agents on machine learning
  experimentation.
\newblock In \emph{International Conference on Machine Learning (ICML)}, 2024.

\bibitem[Hutter et~al.(2011)Hutter, Hoos, and Leyton-Brown]{hutter2011smac}
Frank Hutter, Holger~H. Hoos, and Kevin Leyton-Brown.
\newblock Sequential model-based optimization for general algorithm
  configuration.
\newblock In \emph{Proceedings of LION 2011}, pp.\  507--523, 2011.

\bibitem[Li et~al.(2018)Li, Jamieson, DeSalvo, Rostamizadeh, and
  Talwalkar]{li2017hyperband}
Lisha Li, Kevin Jamieson, Giulia DeSalvo, Afshin Rostamizadeh, and Ameet
  Talwalkar.
\newblock Hyperband: A novel bandit-based approach to hyperparameter
  optimization.
\newblock \emph{Journal of Machine Learning Research}, 18\penalty0
  (185):\penalty0 1--52, 2018.

\bibitem[Liu et~al.(2024{\natexlab{a}})Liu, Gao, and Li]{liu2024agenthpo}
Siyi Liu, Chen Gao, and Yong Li.
\newblock Large language model agent for hyper-parameter optimization.
\newblock \emph{arXiv preprint arXiv:2402.01881}, 2024{\natexlab{a}}.

\bibitem[Liu et~al.(2024{\natexlab{b}})Liu, Astorga, Seedat, and van~der
  Schaar]{liu2024llambo}
Tennison Liu, Nicol{\'a}s Astorga, Nabeel Seedat, and Mihaela van~der Schaar.
\newblock Large language models to enhance bayesian optimization.
\newblock In \emph{International Conference on Learning Representations
  (ICLR)}, 2024{\natexlab{b}}.

\bibitem[Mahammadli \& Ertekin(2024)Mahammadli and
  Ertekin]{mahammadli2024sllmbo}
Kanan Mahammadli and Seyda Ertekin.
\newblock Sequential large language model-based hyper-parameter optimization.
\newblock \emph{arXiv preprint arXiv:2410.20302}, 2024.

\bibitem[Naphade et~al.(2025)Naphade, Bansal, and
  Pareek]{naphade2025smallexpert}
Om~Naphade, Saksham Bansal, and Parikshit Pareek.
\newblock Small {LLMs} with expert blocks are good enough for hyperparameter
  tuning.
\newblock \emph{arXiv preprint arXiv:2509.15561}, 2025.

\bibitem[Olson et~al.(2017)Olson, La~Cava, Orzechowski, Urbanowicz, and
  Moore]{olson2017pmlb}
Randal~S Olson, William La~Cava, Patryk Orzechowski, Ryan~J Urbanowicz, and
  Jason~H Moore.
\newblock Pmlb: A large benchmark suite for machine learning evaluation and
  comparison.
\newblock \emph{BioData Mining}, 10\penalty0 (36), 2017.

\bibitem[O'Malley et~al.(2019)O'Malley, Bursztein, Long, Chollet, Jin,
  Invernizzi, et~al.]{omalley2019kerastuner}
Tom O'Malley, Elie Bursztein, James Long, François Chollet, Haifeng Jin, Luca
  Invernizzi, et~al.
\newblock Keras tuner, 2019.
\newblock URL \url{https://github.com/keras-team/keras-tuner}.

\bibitem[Rychert et~al.(2025)Rychert, Spagnolo, and
  Posashkov]{rychert2025llambo_repro}
Adam Rychert, Gasper Spagnolo, and Evgenii Posashkov.
\newblock Reproducibility study of large language model bayesian optimization.
\newblock \emph{arXiv preprint arXiv:2511.18891}, 2025.

\bibitem[Snoek et~al.(2012)Snoek, Larochelle, and Adams]{snoek2012practical}
Jasper Snoek, Hugo Larochelle, and Ryan~P. Adams.
\newblock Practical bayesian optimization of machine learning algorithms.
\newblock In \emph{Advances in Neural Information Processing Systems},
  volume~25, 2012.

\bibitem[Srinivasan \& Menzies(2026)Srinivasan and
  Menzies]{srinivasan2026beyond}
Srinath Srinivasan and Tim Menzies.
\newblock Beyond the prompt: Assessing domain knowledge strategies for
  high-dimensional hyperparameter optimization.
\newblock \emph{arXiv preprint arXiv:2602.02752}, 2026.

\bibitem[Yang et~al.(2024)Yang, Wang, Lu, Liu, Le, Zhou, and
  Chen]{yang2024opro}
Chengrun Yang, Xuezhi Wang, Yifeng Lu, Hanxiao Liu, Quoc~V Le, Denny Zhou, and
  Xinyun Chen.
\newblock Large language models as optimizers.
\newblock In \emph{International Conference on Learning Representations
  (ICLR)}, 2024.

\bibitem[Zhang et~al.(2023)Zhang, Desai, Bae, Lorraine, and Ba]{liu2023llm_hp}
Michael~R. Zhang, Nishkrit Desai, Juhan Bae, Jonathan Lorraine, and Jimmy Ba.
\newblock Using large language models for hyperparameter optimization, 2023.

\bibitem[Zhang et~al.(2024)Zhang, Yuan, and Avestimehr]{zhang2024revisitopro}
Tuo Zhang, Jinyue Yuan, and Salman Avestimehr.
\newblock Revisiting {OPRO}: The limitations of small-scale {LLMs} as
  optimizers.
\newblock \emph{arXiv preprint arXiv:2405.10276}, 2024.

\end{thebibliography}
\end{document}